\documentclass[accepted]{uai2025} 
                        
\usepackage[american]{babel}
\usepackage{float}
\usepackage{natbib} 
\usepackage{subcaption}
\usepackage{multirow}
\usepackage{balance}
\usepackage{mathtools} 
\usepackage{booktabs} 
\usepackage{tikz} 
\usepackage{amsfonts}

\title{ReReLRP -- Remembering and Recognizing Tasks with LRP}

\author[1,4]{\href{karolina.bogacka.dokt@pw.edu.pl}{Karolina Bogacka}{}}
\author[2]{Maximilian Höfler}
\author[1]{Maria Ganzha}
\author[2]{Wojciech Samek}
\author[3]{Katarzyna Wasielewska-Michniewska}
\affil[1]{%
    Faculty of Mathematics and Information Science\\
    Warsaw University of Technology\\
    Warsaw, Poland
}
\affil[2]{%
    Department of Artificial Intelligence\\
    Fraunhofer Heinrich Hertz Institute\\
    Berlin, Germany
  }
\affil[3]{%
    Systems Research Institute\\
    Polish Academy of Sciences\\
    Warsaw, Poland
  }
\affil[4]{%
    NeverBlink\\
    Warsaw, Poland
}
  
\begin{document}
\maketitle

\begin{abstract}
  Deep neural networks have revolutionized numerous research fields and applications. Despite their widespread success, a fundamental limitation known as catastrophic forgetting remains, where models fail to retain their ability to perform previously learned tasks after being trained on new ones. This limitation is particularly acute in certain continual learning scenarios, where models must integrate the knowledge from new domains with their existing capabilities. Traditional approaches to mitigate this problem typically rely on memory replay mechanisms, storing either original data samples, prototypes, or activation patterns. Although effective, these methods often introduce significant computational overhead, raise privacy concerns, and require the use of dedicated architectures. In this work we present ReReLRP (Remembering and Recognizing with LRP), a novel solution that leverages Layerwise Relevance Propagation (LRP) to preserve information across tasks. Our contribution provides increased privacy of existing replay-free methods while additionally offering built-in explainability, flexibility of model architecture and deployment, and a new mechanism to increase memory storage efficiency. We validate our approach on a wide variety of datasets, demonstrating results comparable with a well-known replay-based method in selected scenarios.
\end{abstract}

\section{Introduction}

In an era where artificial intelligence systems are increasingly integrated into evolving environments, the ability to learn continuously is proving to be essential. Traditional machine learning models, typically trained on static datasets, struggle when new data or tasks are introduced incrementally. Here, tasks are understood as a particular, available to us subset of the problem we would like to train the model to solve over time~\cite{10444954}. They often suffer from~\textbf{catastrophic forgetting}, where learning new information leads to the deterioration of previously acquired knowledge~\cite{French1999catastrophic}. This limitation poses significant challenges for real-world applications such as autonomous driving, personalized healthcare, and adaptive user interfaces, where models must adapt to new information without losing prior expertise.

Continual Learning (CL) addresses this challenge by enabling models to learn from a stream of data over time, effectively retaining performance on old tasks while acquiring new ones~\cite{10444954}. The significance of CL lies in its potential to create systems that are more adaptable, efficient, and closer to human-like learning. 

Particularly in medical applications, the ability to continually learn is crucially important due to the evolving nature of medical data. Access to certain diagnostic categories or imaging modalities can be limited, impairing the quality of the produced model~\cite{Litjens2017survey}. However, the integration of CL methods into real-life scenarios can bring about additional requirements. Due to the particularly sensitive nature of the data, any solution to this problem needs to prioritize \textbf{privacy}. Moreover, \textbf{explainability} is a critical requirement in the medical domain, as practitioners need to understand and trust the decisions made by AI systems~\cite{Samek2017explainable}. Interpretability~\cite{vellido2020importance} not only aids in clinical decision-making, but also ensures compliance with regulatory standards and enhances the acceptance of AI technologies among healthcare professionals.

Despite the progress in CL, many existing class-incremental methods rely on either storing and replaying past data (iCaRL~\cite{Rebuffi2017iCaRL}) or adding complex components to the neural network architecture (DualNet~\cite{pham2021dualnet}). These approaches can be insufficient due to architectural limitations and inadequate capabilities in terms of explainability and privacy. Therefore, there is a pressing need for more flexible and efficient strategies that allow models to retain important knowledge. Recently, NICE~\cite{Gurbuz2024nice} was proposed as one of the first replay-free methods to rival its replay-based contenders in terms of accuracy. However, even though it was inspired by research on explainability, NICE does not support explainable task recognition.

In this paper, we propose a novel method, ReReLRP, that leverages explainable artifical intelligence (XAI) methods to enhance Continual Learning. 
Specifically, we use Layerwise Relevance Propagation (LRP)~\cite{Bach2015lrp} which can be used to interpret neural network predictions by attributing importance scores for each individual neuron. At first, we determine prediction-relevant neurons for each task by computing their relevance scores using LRP. Once these key neurons are identified, we implement a selective freezing mechanism, where the identified neurons are frozen during the training of subsequent tasks. By preventing their weights from being updated, we ensure that the knowledge crucial to previous tasks is preserved.

While the important neurons remain fixed, the unfrozen portions of the network continue to adapt, enabling the model to effectively learn new tasks without interference from the preserved neurons. This approach maintains the plasticity of the model while mitigating catastrophic forgetting. Furthermore,  we incorporate explainable inter-task selection by training inherently interpretable models on the neuron relevance scores computed using LRP. This step adds transparency to the decision-making process, ensuring that the selection of preserved neurons is both data-driven and explainable.

We benchmark ReReLRP against current state-of-the-art methods across various datasets. We also incorporate real-world datasets into our experiments to demonstrate how our approach can translate into practical CL scenarios. Differing network architectures, both in layer composition and size, are included as well. 

 Our contributions involve a novel replay-free method that offers not only privacy and performance competitive to that of replay-based algorithms, similar to NICE, but also inherent explainability, flexibility of deployment, and newly enhanced storage efficiency. Code used to conduct experiments presented in this work is made freely available here: \url{https://github.com/Karolina-Bogacka/ReReLRP}.

\section{Related Works}
Continual Learning (CL), or lifelong learning, enables models to incrementally learn new tasks while preserving prior knowledge. Class-incremental learning~\cite{masana2022class} refers to scenarios, where tasks correspond to subsets of a multi-class classification problem, with the final model integrating knowledge from all tasks.  

Regularization-based methods mitigate forgetting by constraining updates to critical parameters. Elastic Weight Consolidation (EWC)~\cite{Kirkpatrick2017overcoming} penalizes changes to important weights, while Synaptic Intelligence (SI)~\cite{Zenke2017continual} accumulates importance measures to protect significant parameters. Learning Without Forgetting (LWF)~\cite{lwf} employs knowledge distillation to maintain prior knowledge, and Learning Without Memorizing (LWM) enhances this approach by incorporating attention maps.  

Replay-based methods retain past data for training on new tasks. iCaRL~\cite{Rebuffi2017iCaRL} combines exemplar memory with nearest-mean classification, enabling recognition of both old and new classes. Synthetic data generation~\cite{sarmin2024syntheticdatarevisitingprivacyutility} can enhance privacy in replay but remains vulnerable to targeted attacks and requires sufficient training samples~\cite{lu2023machine}.  

Recently, NICE~\cite{Gurbuz2024nice} emerged as a competitive replay-free class-incremental method. It preserves essential neurons by reducing their flexibility and distinguishes tasks using activation patterns. However, its reliance on heavily post-processed activation samples limits interpretability and may discard valuable information.  

Dynamic architecture methods address saturation by expanding network capacity. Progressive Neural Networks~\cite{Rusu2016progressive} add new columns per task, linking them to previous ones for knowledge transfer. Meta-learning approaches, such as meta-experience replay~\cite{Riemer2019learning}, optimize knowledge transfer while reducing interference.  

Explainability-driven methods have also tackled CL. LRP-based approaches~\cite{explain_not_forget} freeze neurons for task-incremental learning but do not address class-incremental challenges. ICICLE~\cite{rymarczyk2023icicle} mitigates interpretability drift, but requires network pretraining and a dedicated architecture, limiting applicability. PiECL~\cite{graphexp} focuses on temporal graph learning, making it less generalizable to broader CL settings.

\section{Preliminaries}

In this section, we provide the necessary background on Continual Learning and Layerwise Relevance Propagation (LRP), which form the foundation of our proposed method.

\subsection{Continual Learning}

Continual Learning (CL) involves training models on a sequence of tasks $\{T_1, T_2, \dots, T_n\}$, where each task $T_k$ is associated with a dataset $\mathcal{D}_k = \{(\mathbf{x}_i^k, y_i^k)\}_{i=1}^{N_k}$~\cite{Parisi2019continual}. The goal is to learn a model that performs well on all observed tasks without accessing data from previous tasks during the training of new tasks.

Let $\theta$ denote the parameters of the neural network model $f(\mathbf{x}; \theta)$. When training on task $T_k$, the objective is to minimize the loss function

\begin{equation}
\mathcal{L}_k(\theta) = \frac{1}{N_k} \sum_{i=1}^{N_k} \ell\left(f(\mathbf{x}_i^k; \theta), y_i^k\right),
\end{equation}

where $\ell(\cdot, \cdot)$ is the loss function, such as cross-entropy loss for classification tasks.

\subsection{Layerwise Relevance Propagation}

Layerwise Relevance Propagation (LRP) is a technique used to interpret the predictions of neural networks by attributing the output of the network back to its input features~\cite{Bach2015lrp}. LRP decomposes the prediction score $f(\mathbf{x})$ of a neural network for an input $\mathbf{x}$ into relevance scores $R_i$ for each input feature $x_i$, such that

\begin{equation}
f(\mathbf{x}) = \sum_i R_i.
\end{equation}

The relevance scores indicate the contribution of each input feature to the prediction. LRP operates by propagating the prediction score backward through the network layers using specific rules that ensure the conservation of relevance at each layer.

For a fully connected neural network, the basic LRP rule for propagating relevance from layer $l+1$ to  $l$ is given by

\begin{equation}
R_i^l = \sum_j \frac{a_i^l w_{ij}^{l,l+1}}{\sum_k a_k^l w_{kj}^{l,l+1} + \varepsilon} R_j^{l+1},
\end{equation}

where $R_i^l$ is the relevance of neuron $i$ at layer $l$, $a_i^l$ is the activation of neuron $i$ at layer $l$, $w_{ij}^{l,l+1}$ is the weight connecting neuron $i$ at layer $l$ to neuron $j$ at layer $l+1$, and $\varepsilon$ is a small constant added for numerical stability.

By applying LRP, we can compute relevance scores for each neuron in the network, identifying which neurons contribute most significantly to the output prediction. 

\section{Proposed Approach}
\begin{figure*}[h!]
\centering
\includegraphics[scale=0.35]{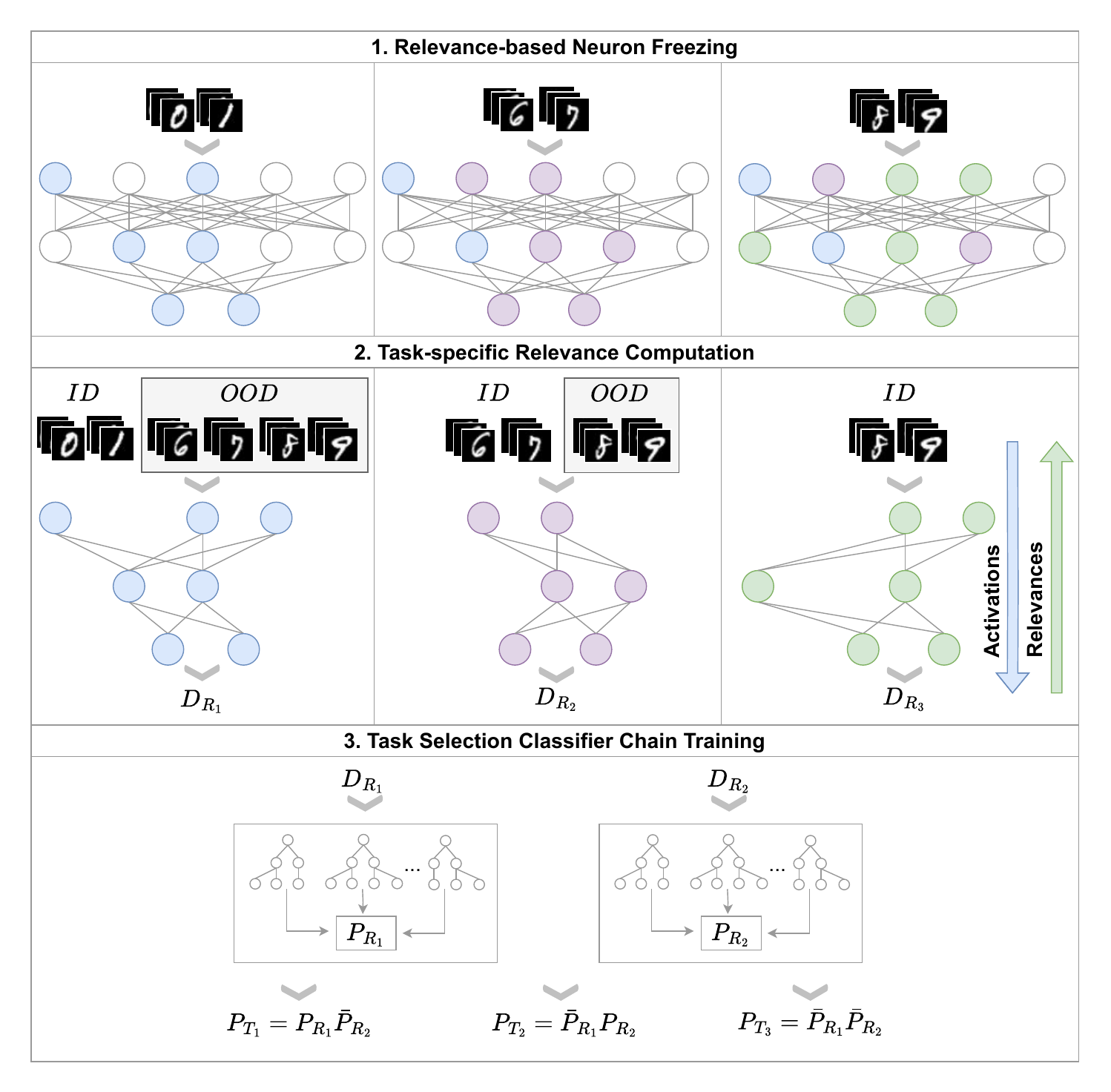}
\caption{A step-by-step illustration of the proposed method. Samples labeled as ID or OOD stand for in-distribution and out-of-distribution, respectively. Relevance datasets built for tasks $1$, $2$, and $3$ are marked as $\mathcal{D}_{R_{1}}$, $\mathcal{D}_{R_{2}}$, and $\mathcal{D}_{R_{3}}$. Probabilities based solely on the relevance subnetworks are expressed as $P_{R_{1}}$ and $P_{R_{2}}$, while $P_{T_{1}}$ and $P_{T_{2}}$ denote probabilities based on all subnetwork relevances. Images used to illustrate the classes are taken from MNIST~\cite{deng2012mnist}.}
\label{fig1}
\end{figure*}

Our approach, termed ReReLRP, is composed of three main stages: Relevance-based Neuron Freezing, Relevance Dataset Construction and Feature Selection, and Task Inference via a Classifier Chain.

\textbf{Relevance-based Neuron Freezing:} Let $\mathcal{M} = (f_\theta, h)$ denote our neural network with parameters $\theta$ and head $h$, which is trained sequentially on tasks $\mathcal{T} = \{T_1, \ldots, T_N\}$. Each task $T_i$ is associated with a dataset $\mathcal{D}_i = \{(x_j^i, y_j^i)\}_{j=1}^{n_i}$. After training on a task, we compute the relevance of each neuron using the mapping function $R_\theta: \mathcal{X} \times \Theta \rightarrow \mathbb{R}^d$. For a neuron $i$ in layer $l$, the relevance is given by 
\[
r_{i,l} = R_\theta(x, \theta_l),
\]
and the absolute relevance scores are normalized per layer as 
\[
\hat{r}_{i,l} = \frac{|r_{i,l}|}{\sum_j |r_{j,l}|}.
\]
Based on these normalized scores, neurons with the smallest average relevance are pruned until a predefined accuracy is achieved, and the remaining (more important) neurons are frozen. Formally, for task $T_i$, the set of frozen neurons is defined as 
\[
\mathcal{F}_i = \{n \in \mathcal{N} : \hat{r}_n > \gamma_i\},
\]
where $\gamma_i$ is chosen so that $\text{Acc}(\mathcal{M}|_{\mathcal{F}_i}, T_i) \geq \tau_i$ for a given accuracy threshold $\tau_i$. This step helps preserve task-specific knowledge and mitigate forgetting.

\textbf{Relevance Dataset Construction and Feature Selection:} To facilitate future task detection, we construct small, targeted relevance datasets
\[
\mathcal{D}_{R_i} = \{(x_j, R_\theta(x_j, \theta))\}_{j=1}^{m},
\]
where the relevance for each input is computed using the predicted class (selected based on the maximal model head output). Representative examples are chosen through a herding strategy, similar to iCaRL~\cite{Rebuffi2017iCaRL}. For feature selection, we consider two approaches. One method involves selecting neurons based on mutual information, denoted by $\mathcal{S}_1 = \{n : I(R_n; T) > \delta\}$ ($I$ standing here for mutual information). Unfortunately, this can incur a high storage overhead. Instead, we opt for a resource-efficient alternative that selects neurons based on the mean relevance over the dataset,
\[
\mathcal{S}_2 = \{n : \mathbb{E}_{x \in \mathcal{D}_i}[|r_n(x)|] > \epsilon\},
\]
which is the method used in all experiments presented here.

\textbf{Task Inference via Classifier Chain:} The final stage leverages the relevance datasets to train a chain of classifiers for task detection. For each task $T_i$, we define its relevance signature as 
\[
S_{T_i}(x) = \{r_n : n \in \mathcal{F}_i\}.
\]
The probability that a sample belongs to task $T_n$ is modeled as follows:
\begin{equation}
\begin{aligned}
p_n = {} & P(T_n | S_{T_n}, \bar{T}_1, \ldots, \bar{T}_{n-1}) \\
& \times \prod_{i=1}^{n-1} \Bigl(1 - P(T_i | S_{T_n}, \bar{T}_1, \ldots, \bar{T}_{i-1})\Bigr),
\end{aligned}
\end{equation}
where $P(T_i | S_{T_i}, \bar{T}_1, \ldots, \bar{T}_{i-1})$ represents the probability output by the classifier chain for task $T_i$, given its relevance signature and the predictions for the preceding tasks.

\section{Experimental Results}
\begin{table*}[t]
\centering
\setlength{\tabcolsep}{2mm} 
\renewcommand{\arraystretch}{1.2} 
\caption{Comparison of different methods on multiple datasets. Accuracy (Acc.) and Forgetting (Forg.) metrics are reported for each dataset. Best results are highlighted in bold.}
\begin{tabular}{l|cc|cc|cc|cc|cc|cc}
    \toprule
    \multirow{2}{*}{} & \multicolumn{2}{c|}{\textbf{EMNIST}} & \multicolumn{2}{c|}{\textbf{FashionMNIST}} & \multicolumn{2}{c|}{\textbf{CIFAR-10}} & \multicolumn{2}{c|}{\textbf{SVHN}} & \multicolumn{2}{c|}{\textbf{CIFAR-100}} & \multicolumn{2}{c}{\textbf{Tiny}}\\
    \midrule
    & \textbf{Acc.} & \textbf{Forg.} & \textbf{Acc.} & \textbf{Forg.} & \textbf{Acc.} & \textbf{Forg.} & \textbf{Acc.} & \textbf{Forg.} & \textbf{Acc.} & \textbf{Forg.} & \textbf{Acc.} & \textbf{Forg.}\\
    \hline
    LWM         & 14.53 & 86.48 & 19.88 & 98.42 & 18.49 & 87.66 & 19.64 & 87.36 & 6.38  & 51.63 & 10.28 & 43.70 \\
    LWF         & 26.47 & 78.06 & 25.17 & 91.88 & 18.97 & 90.90 & 19.39 & 95.64 & 5.95  & 49.40 & \textbf{10.57} & 40.59 \\
    EWC         & 28.83 & 75.74 & 24.49 & 92.68 & 18.81 & 90.10 & 18.97 & 87.44 & 5.65  & 47.60 & 7.85  & 32.75 \\
    iCaRL       & \textbf{57.13} & 14.89 & \textbf{64.04} & 19.45 & 22.51 & 41.82 & \textbf{37.34} & 54.10 & \textbf{10.89} & 11.14 & 0.50  & \textbf{9.10}  \\
    ReReLRP     & 27.05 & \textbf{11.94} & 63.54 & \textbf{14.94} & \textbf{39.91} & \textbf{19.00} & 36.11 & \textbf{21.77} & 5.58  & \textbf{6.03} & 3.33  & 10.49 \\
    \midrule
    \textit{Joint Baseline} & 92.73 & 2.59  & 89.08 & 4.40  & 79.80 & 2.18  & 86.55 & 2.54  & 49.49 & -1.43 & 23.21 & 3.91  \\
    \bottomrule
\end{tabular}
\label{tab:general_results}
\end{table*}

\begin{figure*}[t]
\centering
\begin{subfigure}[b]{0.23\textwidth}
    \centering
    \includegraphics[width=\textwidth]{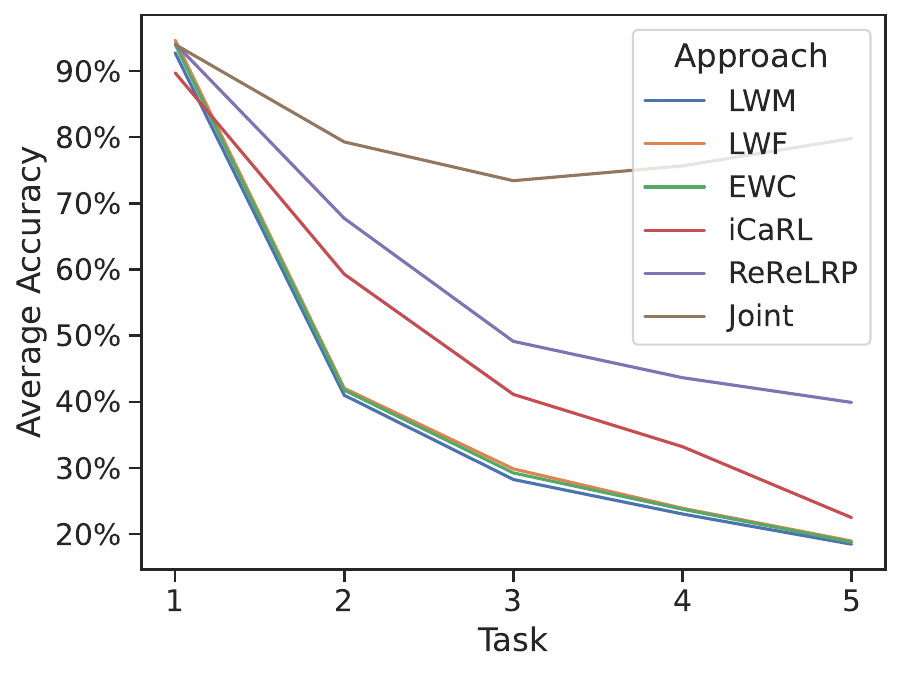}
    \caption{CIFAR10}
    \label{fig:fig2}
\end{subfigure}\hfill
\begin{subfigure}[b]{0.23\textwidth}
    \centering
    \includegraphics[width=\textwidth]{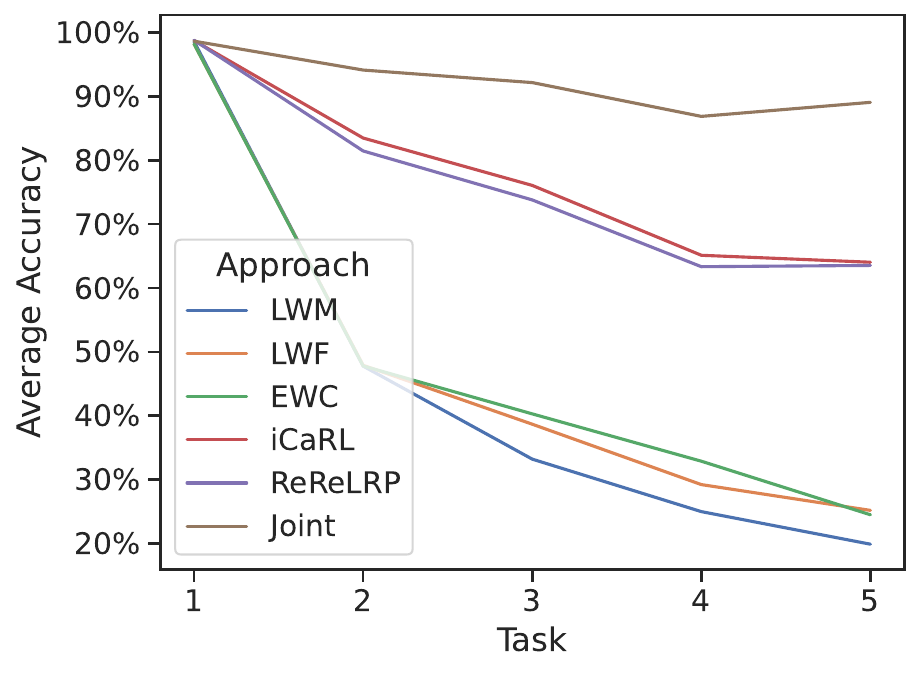}
    \caption{FashionMNIST}
    \label{fig:fig3}
\end{subfigure}\hfill
\begin{subfigure}[b]{0.23\textwidth}
    \centering
    \includegraphics[width=\textwidth]{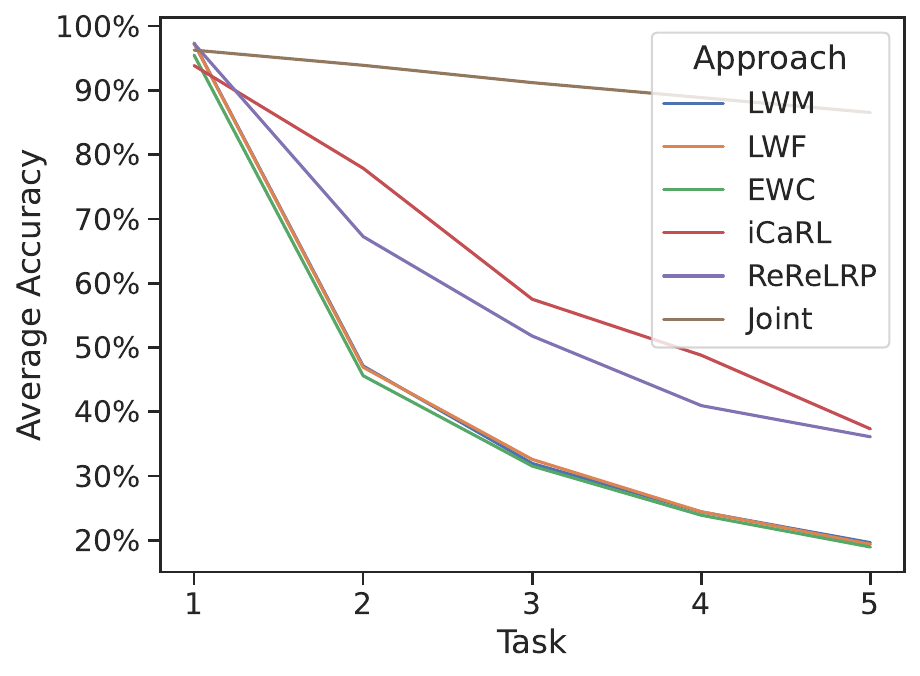}
    \caption{SVHN}
    \label{fig:fig4}
\end{subfigure}\hfill
\begin{subfigure}[b]{0.23\textwidth}
    \centering
    \includegraphics[width=\textwidth]{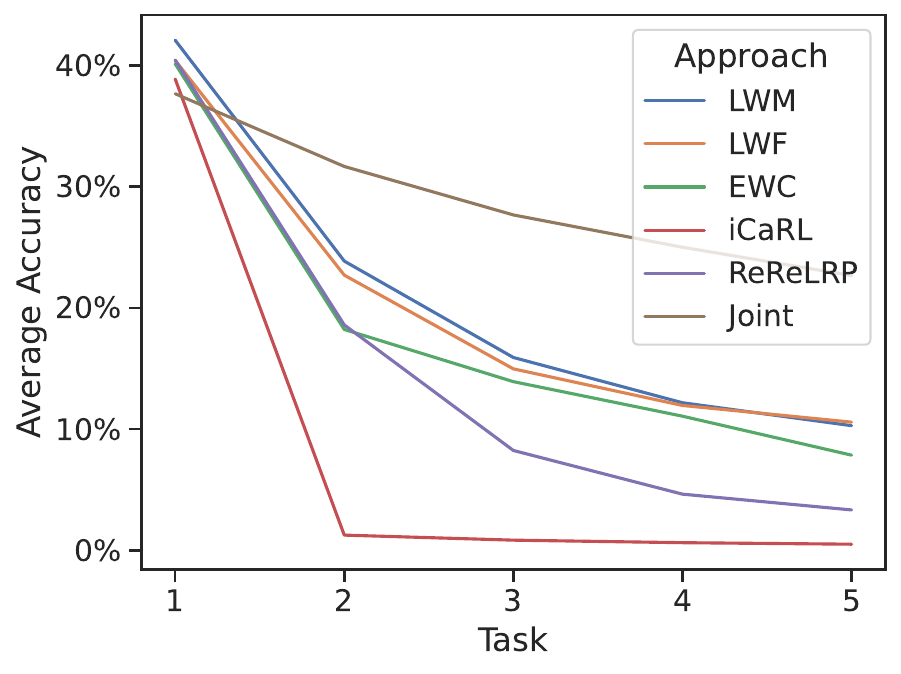}
    \caption{Tiny ImageNet}
    \label{fig:fig5}
\end{subfigure}
\caption{Average task-agnostic accuracy of different approaches.}
\label{fig:cifar_combined}
\end{figure*}
\subsection{Experimental Setup}

\textbf{Random Seeds and Hyperparameter Selection.}  
To ensure the robustness of our results, we conduct all experiments using three different random seeds. A finetuning phase is employed to select the best performing value from three algorithm-specific hyperparameters. Additionally, we conduct experiments for three different initial learning rates and aggregate only the results from the best performing run per seed.

\textbf{Datasets and Model Architectures.}  
To evaluate ReReLRP on a diverse set of model architectures and continual learning tasks, we use different neural network architectures depending on the dataset. Specifically, we utilize LeNet~\cite{lenet} for experiments on EMNIST~\cite{emnist} and FashionMNIST~\cite{fashionmnist}, VGG-11~\cite{vgg} for CIFAR-10 and CIFAR-100~\cite{cifar10}, and ResNet-18~\cite{resnet} for SVHN~\cite{netzer2011reading} and Tiny ImageNet~\cite{tinyimagenet}. The tasks are constructed by following the class order as given by the dataset indexation.

\textbf{Baseline Methods.}  
We benchmark ReReLRP against three well-known replay-free continual learning approaches: Learning Without Forgetting (LWF)~\cite{lwf}, Learning Without Memorizing (LWM)~\cite{lwm}, and Elastic Weight Consolidation (EWC)~\cite{ewc}. To provide a replay-based comparison, we include iCaRL~\cite{Rebuffi2017iCaRL} in our experiments, as well as a joint baseline that is trained on all available task data.

\textbf{Method Implementations and Benchmarking.}  
All methods are implemented using the FACIL framework~\cite{facil}, ensuring a consistent and fair evaluation across different approaches. Further details regarding hyperparameter selection and finetuning strategies are provided in the supplementary material.

\textbf{Evaluation Metrics.}  
We assess model performance using two key metrics: average accuracy and average forgetting, following the benchmark setup of~\cite{derakhshani2022lifelonger}. The average task-aware or task-agnostic accuracy is computed as:
\begin{equation}
    A_{t}=\frac{1}{t}\sum_{i=1}^{t}a_{t,i},
\end{equation}
where $a_{t, i}$ represents the model accuracy on task $i$ after training on task $t$. Additionally, we measure average forgetting, which quantifies the performance drop between the highest and lowest accuracy values for each task:
\begin{equation}
    F=\frac{1}{T-1}\sum_{i=1}^{T-1}\max(a_{t,i}-a_{T,i}).
\end{equation}
These metrics provide a comprehensive understanding of how well the model retains past knowledge while learning new tasks.

\subsection{Comparing to Benchmarks}
\noindent \textbf{Task Performance.} The results shown in Figures~\ref{fig:fig2},~\ref{fig:fig3},~\ref{fig:fig4}, and~\ref{fig:fig5} illustrate the performance of ReReLRP across diverse datasets, highlighting its ability to preserve task-agnostic accuracy and minimize forgetting.

In Figure~\ref{fig:fig2}, ReReLRP achieves particularly high accuracy on CIFAR-10 by effectively freezing task-relevant neurons. This demonstrates its strength in retaining knowledge when the network has sufficient capacity for moderately complex tasks.

Figures~\ref{fig:fig3} and~\ref{fig:fig4} show that ReReLRP maintains competitive performance with iCaRL on FashionMNIST and SVHN while achieving lower forgetting. These results highlight the robustness of relevance-based freezing in adapting to datasets with varying complexity and input characteristics.

In Figure~\ref{fig:fig5}, ReReLRP underperforms on Tiny ImageNet due to network saturation caused by the large number of tasks and the high complexity of the input space. This indicates a limitation of the method when network capacity is insufficient to balance the preservation of prior knowledge with the acquisition of new tasks.

Across all datasets, ReReLRP demonstrates consistent reductions in forgetting, excelling in scenarios where the network has sufficient capacity. Future improvements could focus on dynamic capacity expansion to enhance performance on large-scale datasets with higher task complexity.

\textbf{Overall Performance.} Tables~\ref{tab:general_results} present a comprehensive comparison of average task-agnostic accuracy and forgetting across multiple datasets. Our method, ReReLRP, consistently exhibits notably low forgetting. This trend can be attributed to our aggressive submodel freezing strategy. After the pruning process is complete, the difference between the task-aware accuracy of the preserved submodel and that of the original model on the validation set can be less than 1\% in some configurations, underscoring the effectiveness of our approach in retaining critical task-specific information.

However, the table also reveals a performance trade-off. For example, on EMNIST, while iCaRL achieves a higher accuracy (57.13\%) compared to ReReLRP (27.05\%), ReReLRP achieves slightly lower forgetting (11.94\% vs. 14.89\%). Similar trends are observed on CIFAR-100, where iCaRL outperforms ReReLRP in terms of accuracy, even though ReReLRP records lower forgetting. On the other hand, on datasets like CIFAR-10 and SVHN, ReReLRP not only significantly reduces forgetting (19.00\% and 21.77\%, respectively) but also achieves task-agnostic accuracy competitive or similar to other incremental approaches.

These observations indicate that the benefits of our aggressive submodel freezing are most pronounced in settings where the network has a large number of parameters and is exposed to a relatively low number of tasks. In such scenarios, the redundancy inherent in high-capacity models allows for effective neuron pruning and freezing, leading to robust preservation of past knowledge. Conversely, when applied to architectures with more limited capacity, the aggressive nature of the freezing strategy can impede the acquisition of new knowledge, resulting in lower overall accuracy.

Moreover, while all incremental methods fall short of the joint baseline -- especially in task-agnostic accuracy -- our results highlight that minimizing forgetting is a critical challenge in continual learning, particularly for applications where interpretability and accountability are essential. In such contexts, the ability to attribute decisions to specific, preserved submodels can enhance transparency and trust.

In summary, Table~\ref{tab:general_results} illustrates that while ReReLRP excels in reducing forgetting through aggressive submodel freezing, this benefit is accompanied by a sensitivity to model size and task complexity. Future work will focus on refining the freezing strategy to better balance the trade-off between maintaining past knowledge and learning new tasks, especially in models with limited capacity.

\begin{figure}[t]
\centering

\includegraphics[width=0.3\textwidth]{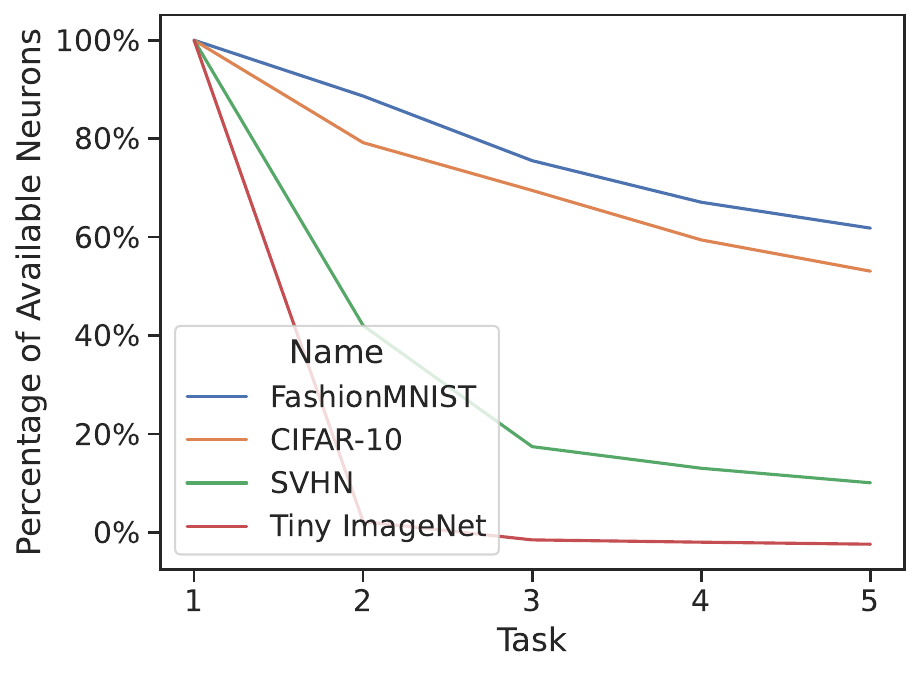} 
\caption{Average percentage of available neurons in experiments conducted on various datasets.}
\label{fig:fig8}

\end{figure}

Figure~\ref{fig:fig8} depicts the percentage of available (that is, unfrozen) neurons after each task for different experiments. Based on the figure and results from Figures~\ref{fig:fig2},~\ref{fig:fig3},~\ref{fig:fig4}, and~\ref{fig:fig5}, it can be observed that ReReLRP is able to maintain competitive performance to other methods when equipped with a large enough network. In experiments conducted on SVHN, CIFAR-10 and FashionMNIST ReReLRP achieves similar or superior accuracy to iCaRL. In Tiny ImageNet, in which a large portion of the network has to be frozen after Task 1, ReReLRP maintains markedly low accuracy. 

\subsection{Memory Requirement}

\begin{table*}[t]
\centering
\setlength{\tabcolsep}{3mm} 
\renewcommand{\arraystretch}{1.2}
\begin{tabular}{l|cc|cc|cc}
    \toprule
    \textbf{Dataset} & \multicolumn{2}{c|}{\textbf{SVHN/CIFAR-10}} & \multicolumn{2}{c|}{\textbf{EMNIST}} & \multicolumn{2}{c}{\textbf{Tiny ImageNet}} \\
    \midrule
    \textbf{Approach} & iCaRL & ReReLRP & iCaRL & ReReLRP & iCaRL & ReReLRP \\
    \midrule
    Number of stored features & \(3\times32\times32\) & \(40\) & \(28\times28\) & \(10\) & \(3\times64\times64\) & \(40\) \\
    Task representatives      & \(5\)            & \(15\) & \(13\)             & \(91\) & \(5\)               & \(15\) \\
    All representatives       & \(5 \times 2\)   & \(15 \times 2\) & \(13 \times 2\)  & \(91 \times 2\) & \(5 \times 40\) & \(15 \times 40\) \\
    Exemplar samples per class & \(6\)           & \(153\) & \(6\)            & \(67\) & \(6\)               & \(614\) \\
    \bottomrule
\end{tabular}
\caption{Comparison of memory budget across different datasets.}
\label{table:memory_budget}
\end{table*}

In continual learning, the memory allocated for storing exemplar representations is critical. In contrast to methods such as iCaRL, which requires storing full exemplar images, our method (ReReLRP) leverages feature preselection to encode a larger number of representative samples while maintaining a comparable memory footprint. Table~\ref{table:memory_budget} presents a memory budget comparison for SVHN, CIFAR-10, EMNIST, and Tiny ImageNet.

For a fair comparison, we enforce the following assumption:
\begin{equation}
\begin{aligned}
n_{f}^{\text{iCaRL}} \times n_{r}^{\text{iCaRL}} \times s^{\text{iCaRL}} \geq 
n_{f}^{\text{ReReLRP}} \\ \times n_{r}^{\text{ReReLRP}} \times s^{\text{ReReLRP}},
\end{aligned}
\end{equation}
where \(n_f\) denotes the number of stored features per exemplar, \(n_r\) represents the number of representative samples per task, and \(s\) is the maximal number of exemplar samples assigned to each class. Here, the number of stored features marks how much space is taken up by a single exemplar. For most of the experiments involving ReReLRP this value has been arbitrarily preselected to be $40$. For experiments on smaller images (e.g., with LeNet), we reduce the size of stored features to $10$ to balance it out with the number of exemplar samples per class. As for task representatives, their number grows faster for ReReLRP than for iCaRL, as relevances for older subnetworks are computed on newer task samples to strengthen task probability estimation. It can be described by the following equation: 
\begin{equation}
n_{r}^{\text{ReReLRP}} = \sum_{i=1}^{t}{i}
\end{equation}
where $t$ indicates the number of task and $n_{r}^{\text{ReReLRP}}$ the number of task representatives needed.
Detailed memory budgets for all experiments are provided in the supplement.

\subsection{Ablation studies}
In order to verify the effect our relevance-based context detector has on the overall performance we include selected ablation experiments. Ablation studies are conducted using the VGG-11 architecture on the CIFAR-10 dataset.

\begin{figure}[!ht]
\centering
\begin{subfigure}[b]{0.23\textwidth}
\centering
\includegraphics[width=\textwidth]{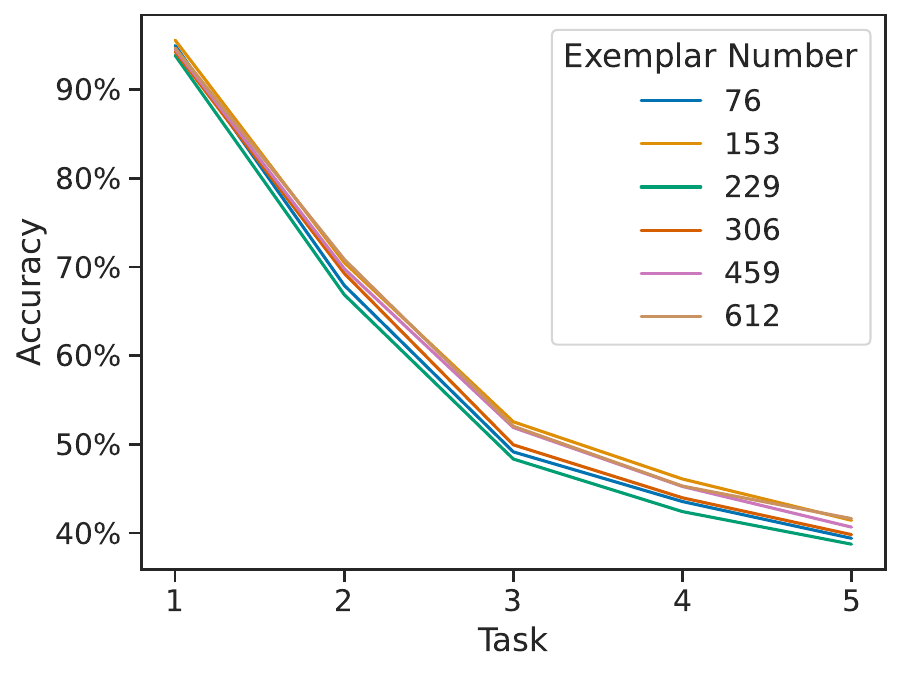} 
\caption{Results for changing exemplar number}
\label{fig:fig6} 
\end{subfigure}
\hfill
\begin{subfigure}[b]{0.23\textwidth}
\includegraphics[width=\textwidth]{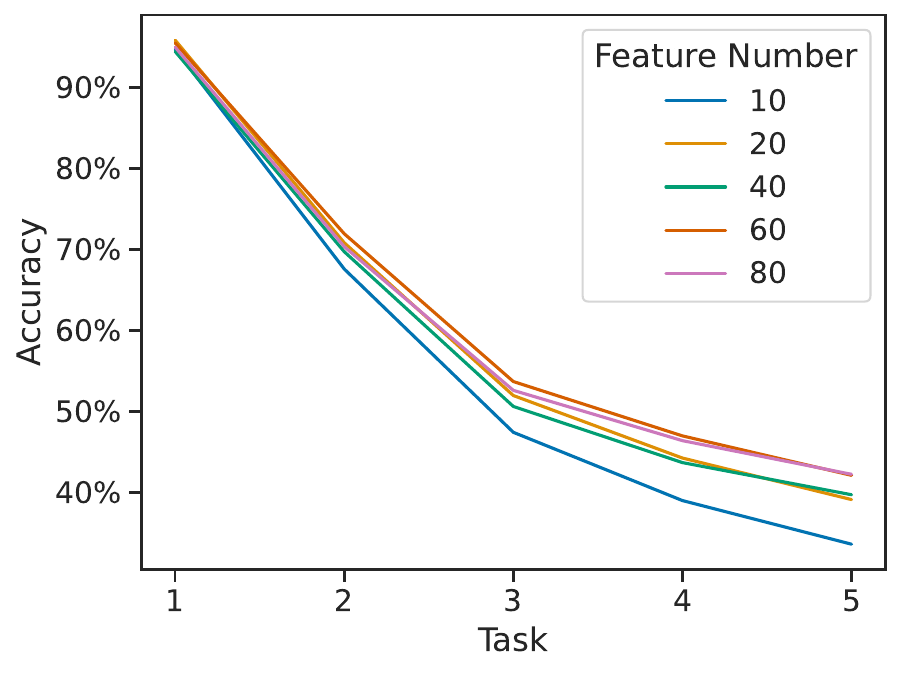} 
\caption{Results for changing feature number}
\label{fig:fig7}
\end{subfigure}
\hfill
\begin{subfigure}[b]{0.23\textwidth}
\includegraphics[width=\textwidth]{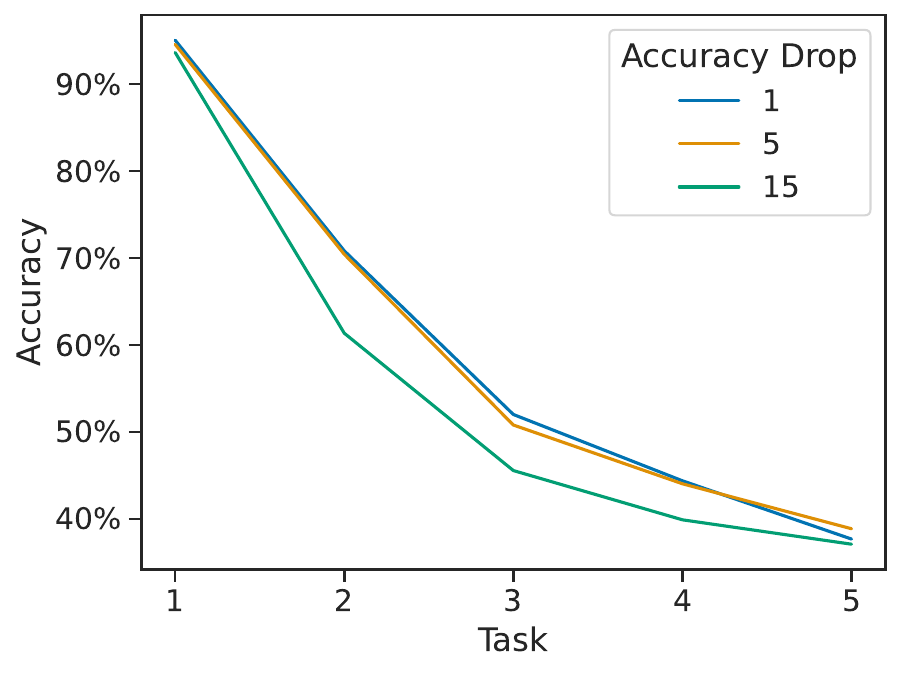} 
\caption{Results for changing accuracy threshold ($\tau$)}
\label{ablfig:fig7}
\end{subfigure}
\label{fig:ablation}
\caption{Average task-agnostic accuracy of ablation experiments.}
\end{figure}
Figures~\ref{fig:fig6},~\ref{fig:fig7}, and~\ref{ablfig:fig7} examine the influence of the number of retained exemplars and features on the task-agnostic accuracy of the method. The results in Figure~\ref{fig:fig6} indicate that increasing the number of preserved exemplars does not consistently lead to higher accuracy. This suggests that beyond a certain threshold, adding more exemplars may not significantly improve the performance, likely due to redundancy in the stored data.

In contrast, Figure~\ref{fig:fig7} demonstrates a clear positive correlation between the number of preserved features and task-agnostic accuracy. Retaining a larger number of features allows the method to better capture the critical task-relevant information, resulting in improved performance over successive tasks. 

Finally, Figure~\ref{ablfig:fig7} investigates the influence of the parameter $\tau$, which indicates the accuracy threshold used during pruning. Although runs with the highest value of $\tau$ achieve markedly lower average accuracy in earlier runs, the difference shrinks with the growing number of tasks. This results suggest a positive relationship between $\tau$ and the ability of the network to adapt to new tasks. Network saturation is the probable culprit behind this result, as a higher value of $\tau$ means more neurons are pruned and available for adaptation after each task.

These findings emphasize the importance of carefully balancing exemplar storage and feature preservation to achieve optimal accuracy while minimizing storage overhead.

\subsection{Results on medical use cases}

To conclude the evaluation, we conduct experiments on a real-world use case relevant to our approach to class-incremental learning. Through the LifeLonger~\cite{derakhshani2022lifelonger} benchmark, we evaluate the performance of ReReLRP on two MedMNIST~\cite{medmnistv2} datasets, BloodMNIST and PathMNIST, with the VGG-11 architecture. We emphasize that medical applications demand robust methods that can handle incremental learning without compromising patient privacy or diagnostic accuracy, making them an ideal testbed for evaluating the practical utility and reliability of continual learning approaches like ReReLRP. Additional details, such as memory budgets and task-aware results, are provided in the supplementary materials.

\begin{figure}[!ht]
\centering
\begin{subfigure}[b]{0.23\textwidth}
    \centering
    \includegraphics[width=\textwidth]{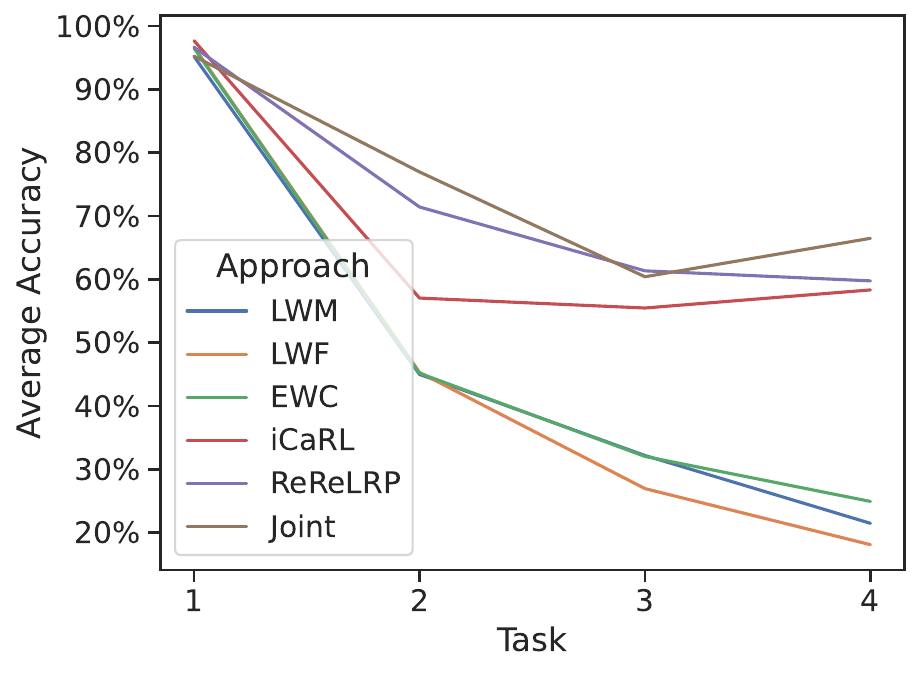}
    \caption{BloodMNIST experiments}
    \label{fig:fig23}
\end{subfigure}\hfill
\begin{subfigure}[b]{0.23\textwidth}
    \centering
    \includegraphics[width=\textwidth]{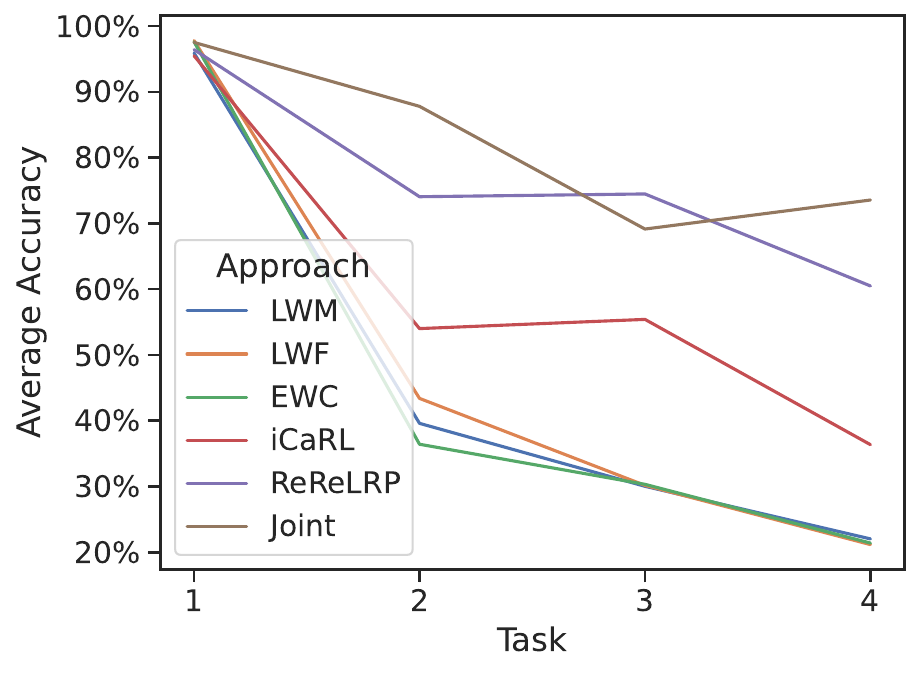}
    \caption{PathMNIST experiments}
    \label{fig:fig24}
\end{subfigure}
\caption{Average task-agnostic accuracy on real-world medical datasets.}
\label{fig:bloodmnist_combined}
\end{figure}

\begin{table}[!ht]
\centering
\setlength{\tabcolsep}{2mm}
\renewcommand{\arraystretch}{1.2} 
\caption{Comparison of different methods on real-world medical datasets. Accuracy (Acc.) and Forgetting (Forg.) metrics are reported for each dataset. Best results are highlighted in bold.}
\begin{tabular}{l|cc|cc}
    \toprule
    \multirow{2}{*}{} & \multicolumn{2}{c}{\textbf{BloodMNIST}} & \multicolumn{2}{c}{\textbf{PathMNIST}} \\
    \midrule
    & \textbf{Acc.} & \textbf{Forg.} & \textbf{Acc.} & \textbf{Forg.} \\
    \hline
    LWM         & 21.46 & 93.89 & 22.03 & 88.39 \\
    LWF         & 18.09 & 89.38 & 21.16 & 90.43 \\
    EWC         & 24.92 & 94.31 & 21.39 & 87.14 \\
    iCaRL       & 58.33 & 17.72 & 36.35 & 22.40 \\
    ReReLRP     & \textbf{59.76} & \textbf{17.37} & \textbf{60.49} & \textbf{19.62} \\
    \midrule
    \textit{Joint Baseline} & 66.48 & 8.25 & 73.55 & 4.60 \\
    \bottomrule
\end{tabular}
\label{tab:real_results}
\end{table}

Figures~\ref{fig:fig23} and~\ref{fig:fig24} present the task-agnostic accuracy of ReReLRP on BloodMNIST and PathMNIST, respectively. These results, in conjunction with Table~\ref{tab:real_results}, demonstrate the competitive performance of ReReLRP in medical image classification tasks. On both datasets, ReReLRP achieves higher accuracy and lower forgetting compared to other methods, except for the joint baseline, which serves as an upper bound. These findings highlight the method's ability to balance task preservation and adaptability, particularly in scenarios with larger models and additional requirements for privacy and explainability.

\section{Discussion}
\textbf{Flexibility and efficient storage}: For continuous functioning, ReReLRP requires storing only a small subset of relevance values computed across the network, enabling efficient task classification without significant storage overhead. By leveraging the task-specific relevance signatures, our method precisely identifies and preserves the critical subnetwork components responsible for modeling each task. This allows for the construction of smaller, task-targeted models that can be easily deployed to perform inference on specific subsets of tasks. This approach could be suitable for resource-constrained environments.

Additionally, ReReLRP does not impose limitations on the architecture of the base model. The relevance-based freezing and masking mechanisms can be applied post hoc to popular network architectures, making the method highly versatile and adaptable. This flexibility ensures that ReReLRP can be integrated into existing models without requiring architectural modifications, while still providing robust task separation and efficient task inference.

\textbf{Explainability through Task-Specific Relevance Signatures}:

ReReLRP leverages layer-wise relevance propagation (LRP) to compute task-specific relevance signatures that play a dual role. By freezing neurons with high relevance after each task, we both preserve critical features to mitigate catastrophic forgetting and generate explicit, interpretable mappings from input features to task decisions. Specifically, for each task \(T_i\), the relevance signature is defined as
\[
S_{T_i}(x) = \{r_n : n \in \mathcal{F}_i\},
\]
which can be projected onto the input space to create detailed visual explanations of the learned features. 

This dual role of relevance -- as a mechanism for preserving knowledge and as a tool for explanation -- renders our approach especially valuable in domains requiring interpretability and accountability, such as medical imaging or autonomous driving. It not only maintains high performance through careful neuron management, but also transforms the model into a transparent system, allowing practitioners to ``see'' what the model is learning at each stage. This transparency enhances both debugging capabilities and user trust, thereby fulfilling a key mandate of explainable AI.

\textbf{Privacy:} ReReLRP eliminates the need to store raw data samples or unprocessed activations from previous tasks, as required by other replay methods, thus significantly reducing potential privacy risks. By relying solely on relevance values and task-specific subnetwork structures, our method ensures that sensitive user data is not retained, thereby enhancing compliance with privacy-preserving requirements in real-world applications, including medical settings.

\textbf{Limitations:} The current approach of freezing neurons after each task consumes a substantial portion of the neural network's parameter space. As the number of tasks increases, the fixed capacity of the network can lead to saturation, ultimately limiting the method's scalability and effectiveness in large-scale, long-term learning scenarios. Moreover, the reliance on aggressive neuron freezing may reduce flexibility when adapting to new tasks with overlapping feature requirements. Further research is necessary to explore strategies mitigating these limitations while preserving task performance.

\textbf{Outlook:} Future work could investigate advanced feature selection techniques that reduce the storage of relevance values while maintaining high inter-task classification performance. Such improvements could enhance the efficiency of the method and reduce its memory overhead, particularly in memory-constrained applications. Another promising direction is the integration of dynamic network capacity management, allowing the network to expand incrementally as new tasks are introduced. Additionally, the concept of relevance scores could be extended to enhance both explainability and performance in class-incremental learning scenarios. By repurposing these scores, stored data could serve dual purposes: improving interpretability through explainable AI techniques and facilitating effective continual learning. Furthermore, applying this method to scenarios with large image sizes relative to network capacity could provide a scalable and storage-efficient solution for real-world applications.

\section{Conclusions}

In this work, we introduce ReReLRP, a novel method for class-incremental learning that combines layer-wise relevance propagation (LRP) with explainable task-specific neuron freezing. To the best of our knowledge, this is the first approach to utilize LRP metrics for not only task performance preservation, but also explainable task recognition in the context of continual learning. Our experiments demonstrate that ReReLRP achieves performance comparable to state-of-the-art techniques across a variety of datasets while offering significant advantages in explainability and memory efficiency. These results highlight the potential of relevance-driven approaches to address key challenges in class-incremental learning, paving the way for future advancements in this area.
\balance
\begin{contributions} 
    Karolina Bogacka conceived the idea, created the code and wrote the majority of the paper,
    Maximilian Höfler was responsible for theoretical consultations, a large part of the bibliography, mathematical notation and explanation of Layerwise Relevance Propagation,
    Wojciech Samek, Maria Ganzha, and Katarzyna Wasielewska-Michniewska provided feedback on the experiment methodology and the contents of the paper.
\end{contributions}

\begin{acknowledgements} 
    The work of Karolina Bogacka was conducted during a research visit funded by the Warsaw University of Technology within the Excellence Initiative: Research University(IDUB) programme.
\end{acknowledgements}

\bibliography{uai}

\newpage

\onecolumn

\title{ReReLRP -- Remembering and Recognizing Tasks with LRP\\(Supplementary Material)}
\maketitle

This Supplementary Material should be submitted together with the main paper.

\appendix
\section{Additional exemplar budgets}
Tables~\ref{table10} and~\ref{table11} list exemplar budgets computed for other benchmark experiments. Tables~\ref{table12} and~\ref{table13} list exemplar budgets computed for real-life use case experiments. 

\begin{table}[!ht]
\centering
\setlength{\tabcolsep}{1mm} 

\begin{tabular}{l|l|l}
    \textbf{Approach} & \textbf{iCaRL} & \textbf{ReReLRP}  \\
    \hline
    Number of stored features & $28\cdot28$ & $10$ \\
    Task representatives & $5$ & $5+4+3+2+1$ \\
    All representatives & $5*2$ & $15*2$ \\
    Exemplar samples per class & $6$ & $156$ \\
\end{tabular}

\caption{Memory budget for FashionMNIST experiments.}
\label{table10}
\end{table}

\begin{table}[!ht]
\centering
\setlength{\tabcolsep}{1mm} 

\begin{tabular}{l|l|l}
    \textbf{Approach} & \textbf{iCaRL} & \textbf{ReReLRP}  \\
    \hline
    Number of stored features & $3\cdot28\cdot28$ & $40$ \\
    Task representatives & $10$ & $55$ \\
    All representatives & $100$ & $550$\\
    Exemplar samples per class & $6$ & $83$ \\
\end{tabular}

\caption{Memory budget for CIFAR-100 experiments.}
\label{table11}
\end{table}

\begin{table}[!ht]
\centering
\begin{tabular}{l|l|l}
    \textbf{Approach} & \textbf{iCaRL} & \textbf{ReReLRP}  \\
    \hline
    Number of stored features & $28\cdot28$ & $40$ \\
    Task representatives & $4$ & $4+3+2+1$ \\
    All representatives & $4*2$ & $10*2$ \\
    Exemplar samples per class & $6$ & $141$ \\
\end{tabular}

\caption{Memory budget for BloodMNIST experiments.}
\label{table12}
\end{table}

\begin{table}[!ht]
\centering
\setlength{\tabcolsep}{1mm} 

\begin{tabular}{l|l|l}
    \textbf{Approach} & \textbf{iCaRL} & \textbf{ReReLRP}  \\
    \hline
    Number of stored features & $3\cdot28\cdot28$ & $40$ \\
    Task representatives & $4$ & $4+3+2+1$ \\
    All representatives & $3+3*2$ & $4*3+6*2$\\
    Exemplar samples per class & $6$ & $132$ \\
\end{tabular}

\caption{Memory budget for PathMNIST experiments.}
\label{table13}
\end{table}

\section{Additional experiment details}

\subsection{Benchmark experiments}
The specific starting learning rates used are $0.1$, $0.05$, and $0.01$. 
The method-specific parameter values tried during the finetuning phase are as follows:
\begin{itemize}
  \item EWC---$\lambda \in \{3000, 5000, 7000\}$
  \item LWF---$\lambda \in \{0.5, 1, 2\}$
  \item LWM---$\beta \in \{0.5, 1, 2\}$
  \item iCaRL---$\lambda \in \{0.5, 1, 2\}$
  \item ReReLRP---$\tau \in \{1, 5, 15\}$
\end{itemize}
The values for each method were selected based on their default setting in FACIL~\cite{facil}, so that both a more flexible and more rigid setting than the default is incorporated into the experiments.

The best performing runs for each seed are judged by their final average task-agnostic accuracy. Similar approach is used to select the most suitable parameter during the finetuning phase.

To ensure fair evaluation and adhere to the benchmark assumptions, we add a knowledge distillation loss term and include the same number of exemplars as assigned to iCaRL in LWF, LWM, and EWC examples. We use herding for exemplar selection in all cases. For experiments conducted on ResNet-18, we freeze the batch norm layers after the first task. This is done to ensure a fair comparison between methods employing neuron freezing and other methods.

For explainable task selection probability modeling, we employ the random forest algorithm~\cite{parmar2019review}. The ablation experiments were computed as an average over $3$ different seed values as well.

\subsection{Experiments on real-life use cases}

The learning rate, as well as the finetuning configuration in this set of experiments is the same as in the benchmark experiments. Additionally, the order of class appearance was kept identical to the order of class numbers. Referring to class division into tasks, BloodMNIST is divided into $4$ tasks containing $2$ classes each. PathMNIST starts with a task consisting of $3$ classes, followed by $3$ tasks with $2$ classes each---$4$ tasks in total. 

\section{Additional plots illustrating network saturation} 

\begin{figure*}[!ht]
\centering
\begin{subfigure}[b]{0.4\textwidth}
    \centering
    \includegraphics[width=\textwidth]{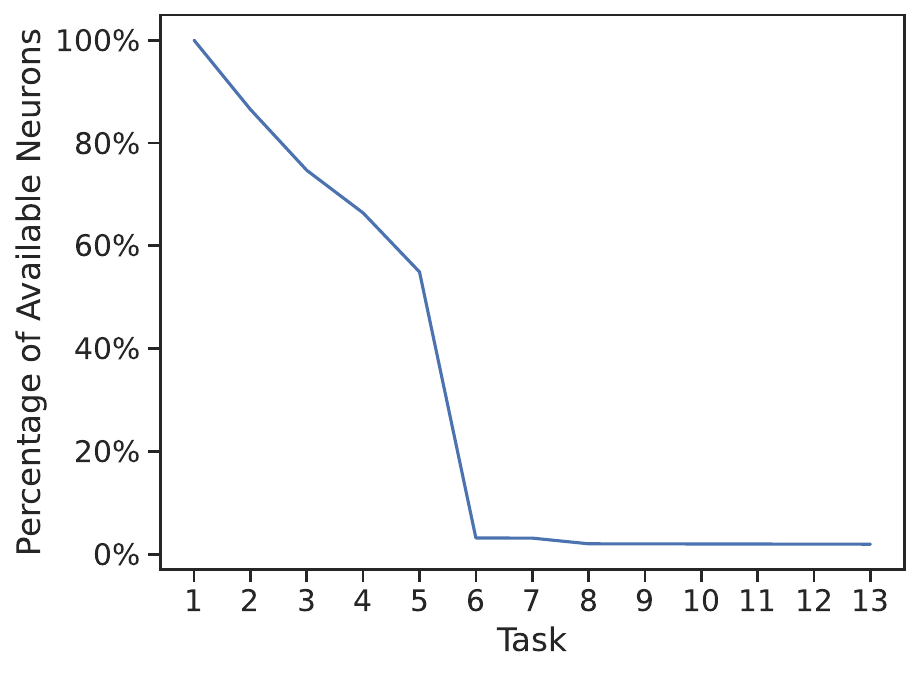}
    \caption{EMNIST experiments}
    \label{fig:fig21} 
\end{subfigure}
\hspace{0.05\textwidth}
\begin{subfigure}[b]{0.4\textwidth}
    \centering
    \includegraphics[width=\textwidth]{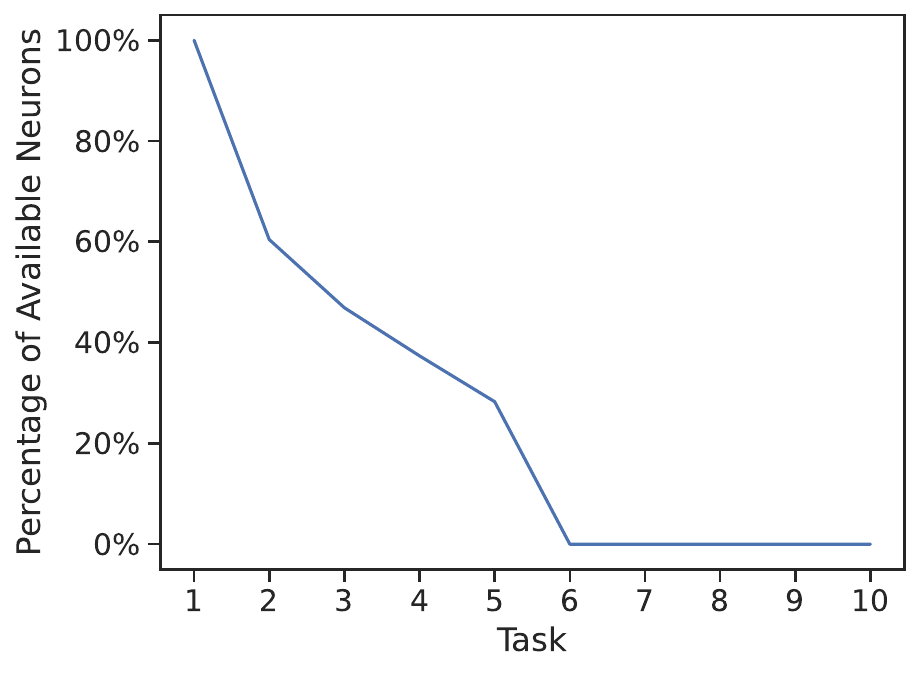}
    \caption{CIFAR-100 experiments}
    \label{fig:fig22} 
\end{subfigure}
\label{fig:pathmnist_combined}
\caption{Network saturation expressed in the percentage of available (unfrozen) neurons.}
\end{figure*}

Figures~\ref{fig:fig21} and~\ref{fig:fig22} demonstrate the network saturation in CIFAR-100 and EMNIST experiments. In both, the network becomes fully saturated in the middle of their runs, which leads to significantly inferior performance.

\section{Additional experiment results} 
\subsection{Benchmark experiments}
\begin{figure*}[!ht]
\centering
\begin{subfigure}[b]{0.4\textwidth}
    \centering
    \includegraphics[width=\textwidth]{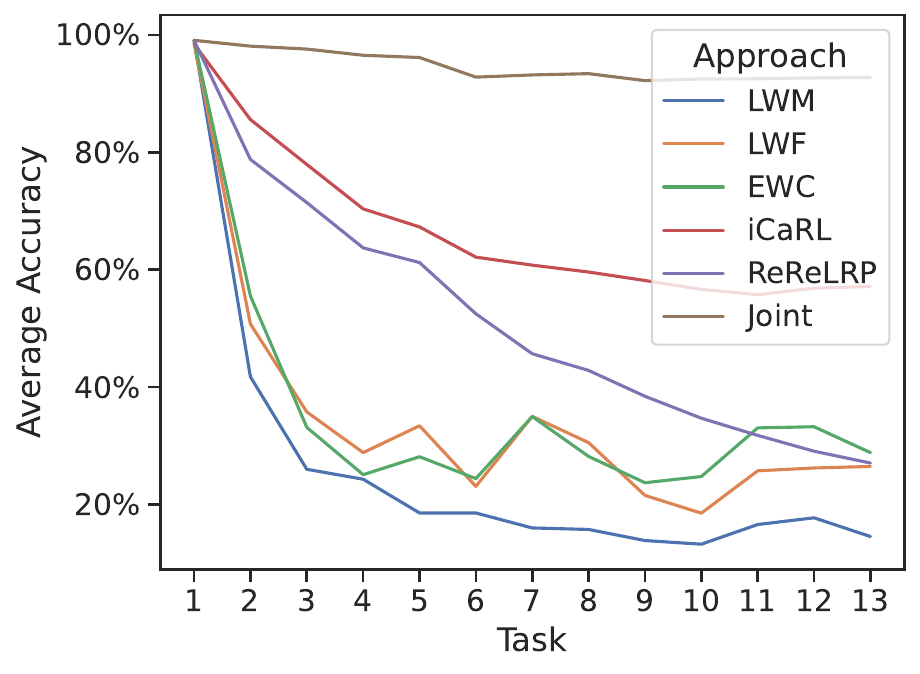}
    \caption{Average task-agnostic accuracy}
    \label{fig:fig13} 
\end{subfigure}
\hspace{0.05\textwidth}
\begin{subfigure}[b]{0.4\textwidth}
    \centering
    \includegraphics[width=\textwidth]{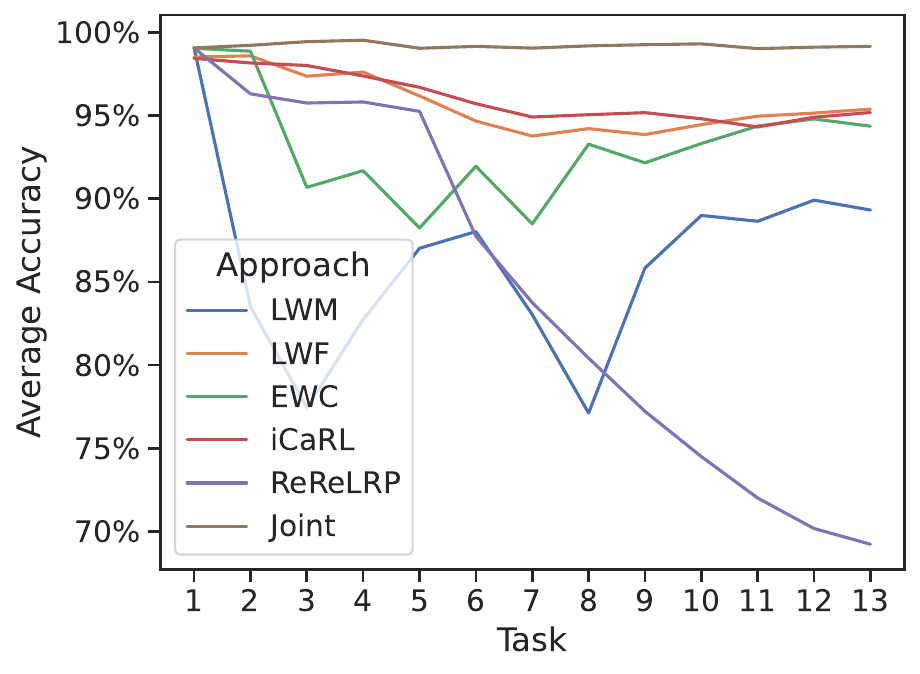}
    \caption{Average task-aware accuracy}
    \label{fig:fig14} 
\end{subfigure}
\caption{Performance of different approaches measured on the EMNIST dataset.}
\label{fig:emnist_combined}
\end{figure*}

\begin{figure*}[!ht]
\centering
\begin{subfigure}[b]{0.4\textwidth}
    \centering
    \includegraphics[width=\textwidth]{images/svhn}
    \caption{Average task-agnostic accuracy}
    \label{fig:fig15} 
\end{subfigure}
\hspace{0.05\textwidth}
\begin{subfigure}[b]{0.4\textwidth}
    \centering
    \includegraphics[width=\textwidth]{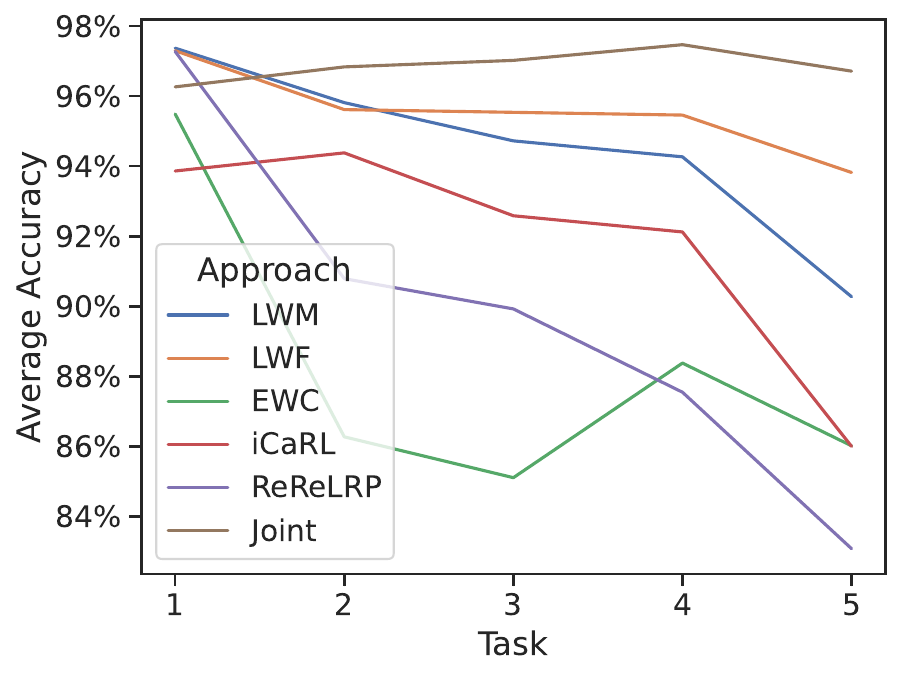}
    \caption{Average task-aware accuracy}
    \label{fig:fig16} 
\end{subfigure}
\caption{Performance of different approaches measured on the SVHN dataset.}
\label{fig:svhn_combined}
\end{figure*}

\begin{figure*}[!ht]
\centering
\begin{subfigure}[b]{0.4\textwidth}
    \centering
    \includegraphics[width=\textwidth]{images/tiny}
    \caption{Average task-agnostic accuracy}
    \label{fig:fig17} 
\end{subfigure}
\hspace{0.05\textwidth}
\begin{subfigure}[b]{0.4\textwidth}
    \centering
    \includegraphics[width=\textwidth]{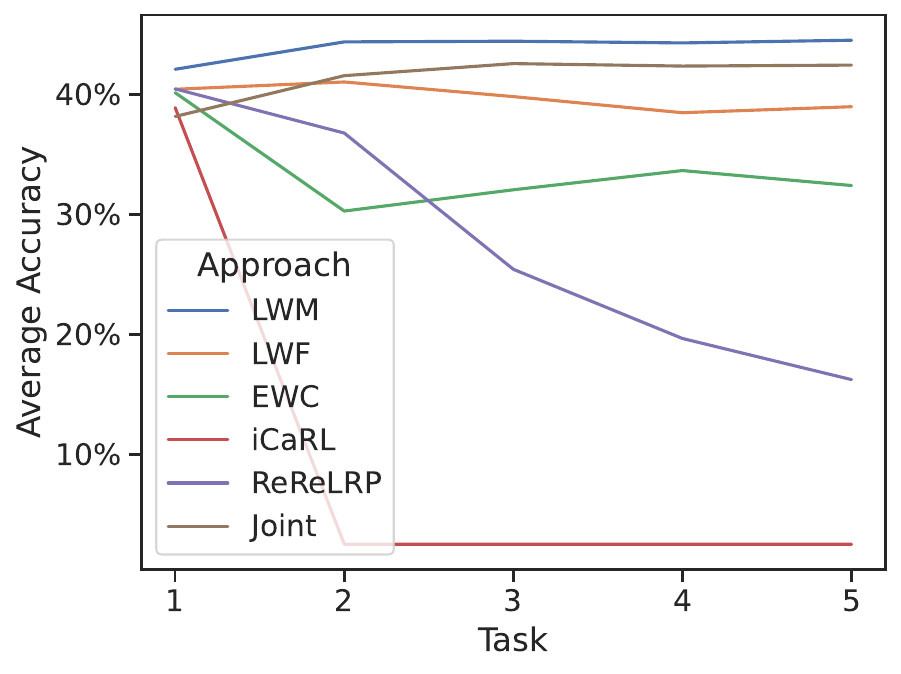}
    \caption{Average task-aware accuracy}
    \label{fig:fig18} 
\end{subfigure}
\caption{Performance of different approaches measured on the Tiny ImageNet dataset.}
\label{fig:tiny_combined}
\end{figure*}

\begin{figure*}[!ht]
\centering
\begin{subfigure}[b]{0.4\textwidth}
    \centering
    \includegraphics[width=\textwidth]{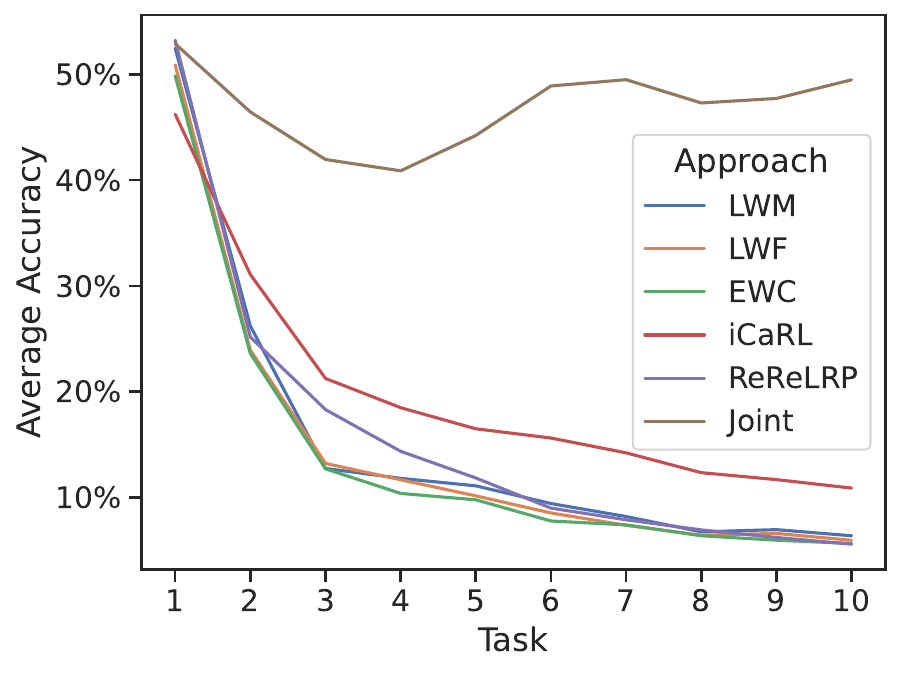}
    \caption{Average task-agnostic accuracy}
    \label{fig:fig19} 
\end{subfigure}
\hspace{0.05\textwidth}
\begin{subfigure}[b]{0.4\textwidth}
    \centering
    \includegraphics[width=\textwidth]{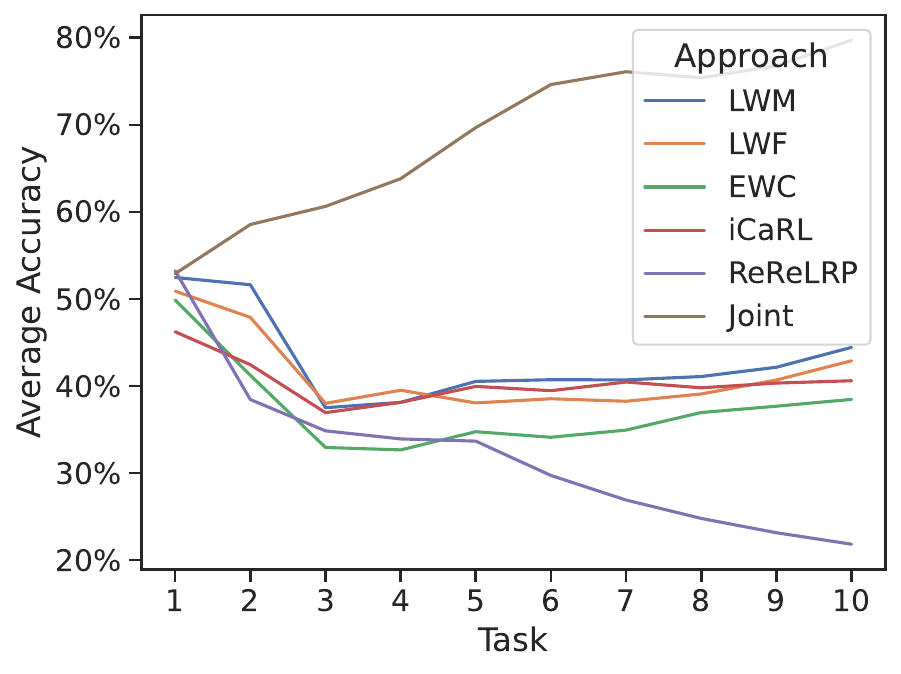}
    \caption{Average task-aware accuracy}
    \label{fig:fig20} 
\end{subfigure}
\caption{Performance of different approaches measured on the CIFAR-100 dataset.}
\label{fig:cifar100_combined}
\end{figure*}

\begin{figure*}[!ht]
\centering
\begin{subfigure}[b]{0.4\textwidth}
    \centering
    \includegraphics[width=\textwidth]{images/cifar10}
    \caption{Average task-agnostic accuracy}
    \label{fig:fig30} 
\end{subfigure}
\hspace{0.05\textwidth}
\begin{subfigure}[b]{0.4\textwidth}
    \centering
    \includegraphics[width=\textwidth]{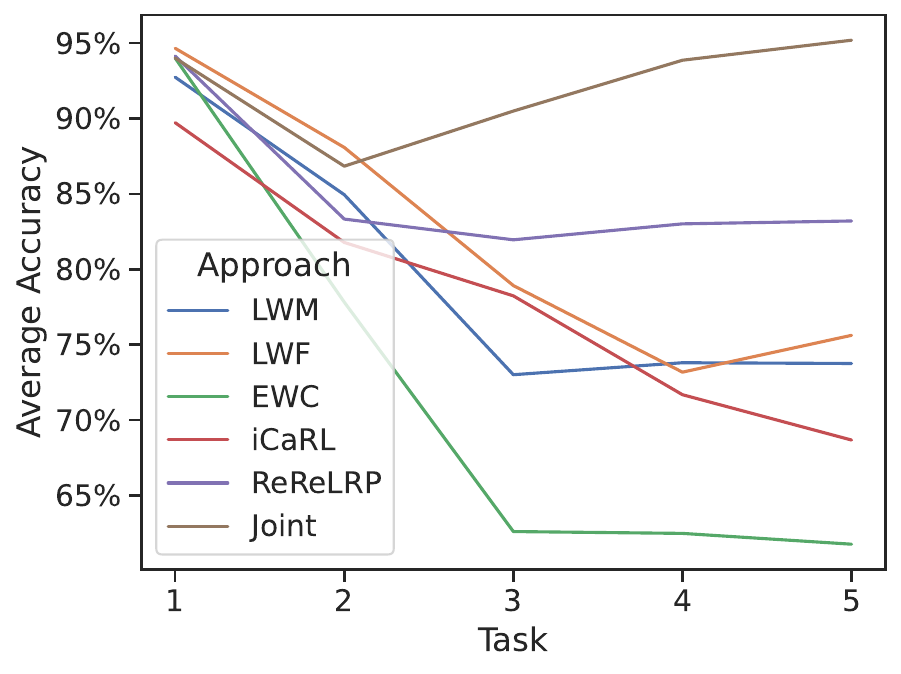}
    \caption{Average task-aware accuracy}
    \label{fig:fig31} 
\end{subfigure}
\caption{Performance of different approaches measured on the CIFAR-10 dataset.}
\label{fig:cifar10_combined}
\end{figure*}

\begin{figure*}[!ht]
\centering
\begin{subfigure}[b]{0.4\textwidth}
    \centering
    \includegraphics[width=\textwidth]{images/fashionmnist.pdf}
    \caption{Average task-agnostic accuracy}
    \label{fig:fig30} 
\end{subfigure}
\hspace{0.05\textwidth}
\begin{subfigure}[b]{0.4\textwidth}
    \centering
    \includegraphics[width=\textwidth]{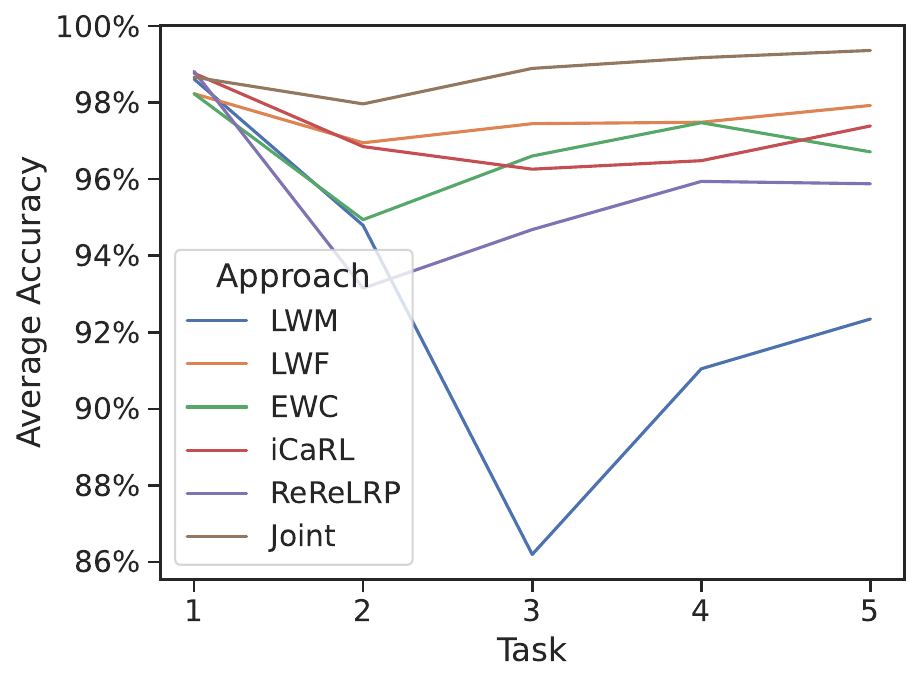}
    \caption{Average task-aware accuracy}
    \label{fig:fig31} 
\end{subfigure}
\caption{Performance of different approaches measured on the FashionMNIST dataset.}
\label{fig:fashionmnist_combined}
\end{figure*}

Figures~\ref{fig:emnist_combined},~\ref{fig:svhn_combined},~\ref{fig:tiny_combined},~\ref{fig:cifar100_combined},~\ref{fig:cifar10_combined}, and~\ref{fig:fashionmnist_combined} contain full results for experiments conducted on EMNIST, SVHN, Tiny ImageNet, CIFAR-100, CIFAR-10, and FashionMNIST respectively. Each figure contains two subfigures---one visualizing the average task-agnostic accuracy, the other the average task-aware accuracy. Table~\ref{table12} extends the results presented in the main paper in Table~\ref{tab:general_results} to task-aware measurements.

\begin{table*}[!ht]
\centering
\setlength{\tabcolsep}{1mm} 
\begin{tabular}{l|l|l|l|l|l|l|l|l|l|l|l|l}
    
    & \multicolumn{2}{|l|}{\textbf{EMNIST}} & \multicolumn{2}{|l}{\textbf{FashionMNIST}} & \multicolumn{2}{|l}{\textbf{CIFAR-10}} & \multicolumn{2}{|l}{\textbf{SVHN}} & \multicolumn{2}{|l}{\textbf{CIFAR-100}} & \multicolumn{2}{|l}{\textbf{Tiny ImageNet}}\\
    \hline
    & Acc. & Forg. & Acc. & Forg. & Acc. & Forg. & Acc. & Forg. & Acc. & Forg. & Acc. & Forg.\\
    \hline
    LWM & $89.32$ & $7.54$ & $92.35$ & $7.85$ & $73.75$ & $18.58$ & $90.27$ & $3.25$ & $\textbf{44.44}$ & $9.36$ & $\textbf{44.51}$ & $1.19$\\
    LWF & $\textbf{95.38}$ & $3.47$ & $\textbf{97.93}$ & $0.97$ & $75.62$ & $20.09$ & $\textbf{93.82}$ & $2.72$ & $42.90$ & $8.35$ & $38.96$ & $5.17$\\
    EWC & $94.36$ & $4.78$ & $96.71$ & $2.41$ & $61.77$ & $36.40$ & $86.02$ & $4.81$ & $38.47$ & $11.13$ & $32.39$ & $5.15$\\
    iCaRL & $95.18$ & $1.67$ & $97.39$ & $1.80$ & $68.67$ & $19.29$ & $86.00$ & $11.34$ & $40.61$ & $2.93$ & $2.50$ & $9.10$\\
    ReReLRP & $69.23$ & $\textbf{0.18}$ & $95.88$ & $\textbf{0.76}$ & $\textbf{83.20}$ & $\textbf{1.09}$ & $82.32$ & $\textbf{0.86}$ & $21.84$ & $\textbf{1.14}$ & $16.22$ & $\textbf{0.31}$\\\cline{1-13}
    Joint baseline & $99.16$ & $0.07$ & $99.36$ & $-0.07$ & $95.19$ & $-1.00$ & $96.71$ & $-0.11$ & $79.72$ & $-2.48$ & $42.43$ & $-0.84$\\
\end{tabular}

\caption{Method performance measured on different datasets. Accuracy (Acc.) and Forgetting (Forg.) metrics are reported for each dataset. Best results are highlighted in bold.}
\label{table12}
\end{table*}
\subsection{Experiments on real-life use cases}

\begin{figure*}[!ht]
\centering
\begin{subfigure}[b]{0.4\textwidth}
    \centering
    \includegraphics[width=\textwidth]{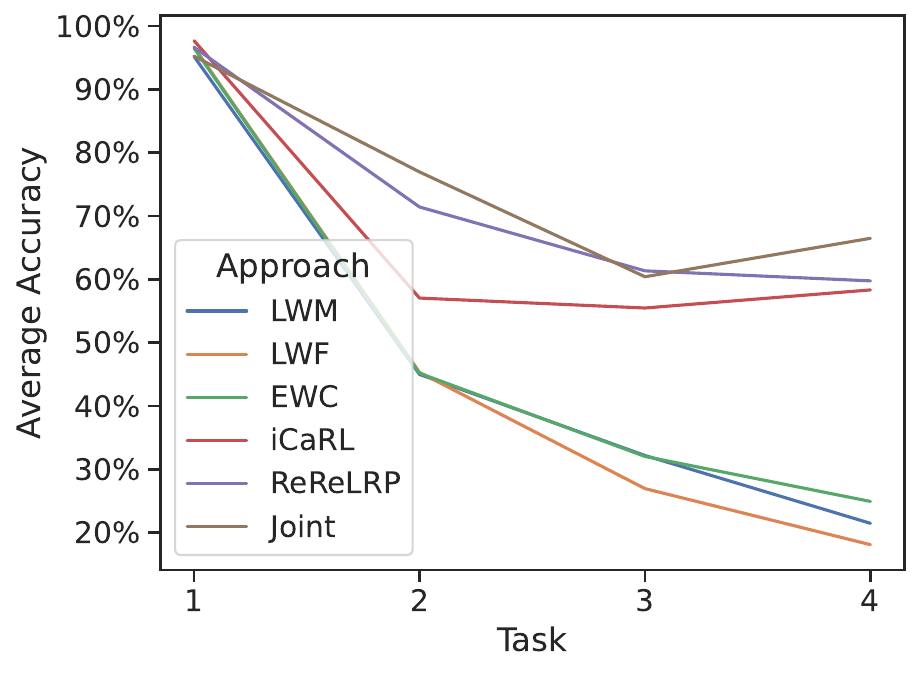}
    \caption{Average task-agnostic accuracy}
    \label{fig:fig30} 
\end{subfigure}
\hspace{0.05\textwidth}
\begin{subfigure}[b]{0.4\textwidth}
    \centering
    \includegraphics[width=\textwidth]{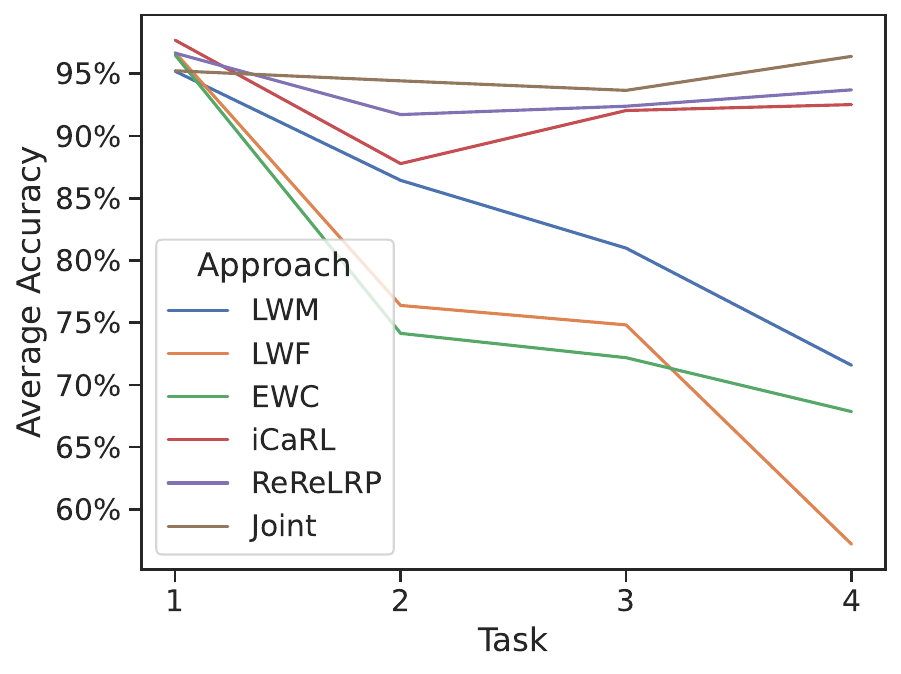}
    \caption{Average task-aware accuracy}
    \label{fig:fig31} 
\end{subfigure}
\caption{Performance of different approaches measured on the BloodMNIST dataset.}
\label{fig:blood_combined2}
\end{figure*}

\begin{figure*}[!ht]
\centering
\begin{subfigure}[b]{0.4\textwidth}
    \centering
    \includegraphics[width=\textwidth]{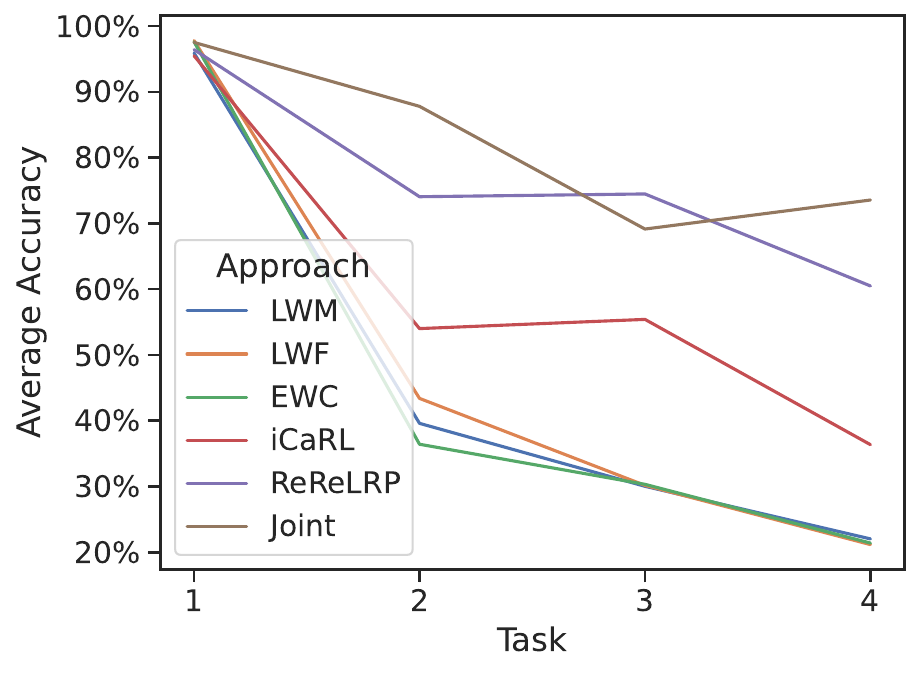}
    \caption{Average task-agnostic accuracy}
    \label{fig:fig30} 
\end{subfigure}
\hspace{0.05\textwidth}
\begin{subfigure}[b]{0.4\textwidth}
    \centering
    \includegraphics[width=\textwidth]{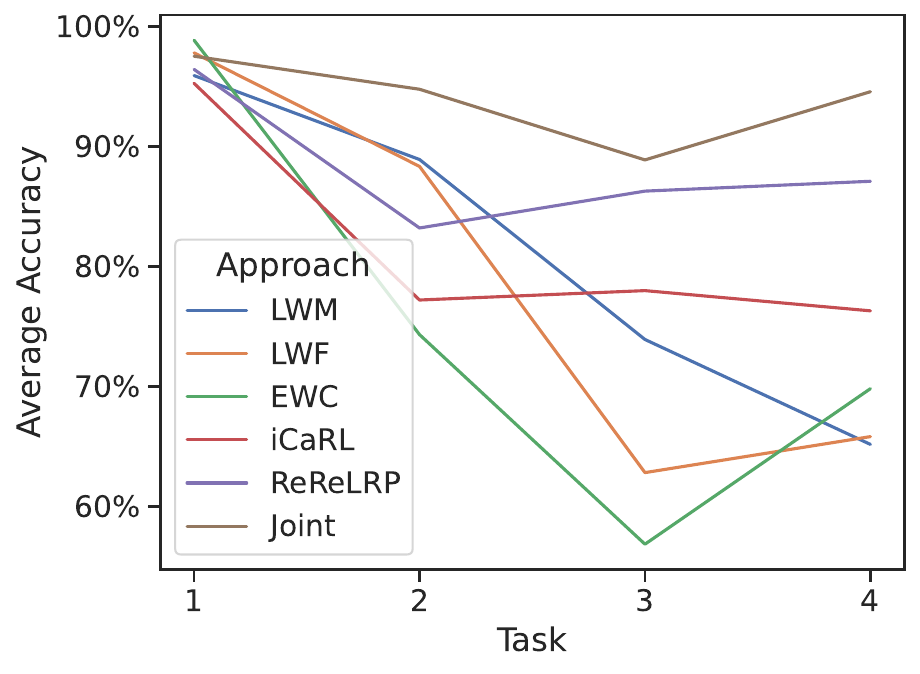}
    \caption{Average task-aware accuracy}
    \label{fig:fig31} 
\end{subfigure}
\caption{Performance of different approaches measured on the PathMNIST dataset.}
\label{fig:pathcombined2}
\end{figure*}

\begin{table}[!ht]
\centering
\setlength{\tabcolsep}{2mm} 
\renewcommand{\arraystretch}{1.2} 
\caption{Comparison of task-aware performance of different methods on real-life medical datasets. Accuracy (Acc.) and Forgetting (Forg.) metrics are reported for each dataset. Best results are highlighted in bold.}
\begin{tabular}{l|cc|cc}
    \toprule
    \multirow{2}{*}{} & \multicolumn{2}{c|}{\textbf{BloodMNIST}} & \multicolumn{2}{c|}{\textbf{PathMNIST}}\\
    \midrule
    & \textbf{Acc.} & \textbf{Forg.} & \textbf{Acc.} & \textbf{Forg.} \\
    \hline
    LWM         & 71.58 & 27.21 & 65.19 & 32.81  \\
    LWF         & 57.21 & 37.22 & 65.84 & 30.86  \\
    EWC         & 67.86 & 37.06 & 69.83 & 22.50  \\
    iCaRL       & 92.52 & 3.37 & 76.31 & 4.87  \\
    ReReLRP     & \textbf{93.71} & \textbf{0.47} & \textbf{87.10} & \textbf{3.13} \\
    \midrule
    \textit{Joint Baseline} & 96.40 & -1.02  & 94.56 & -2.08 \\
    \bottomrule
\end{tabular}

\label{tab:real_results_taw}
\end{table}

Figures~\ref{fig:blood_combined2} and~\ref{fig:pathcombined2} contain full results for experiments conducted on BloodMNIST and PathMNIST respectively. Each figure contains two subfigures---one visualizing the average task-agnostic accuracy, the other the average task-aware accuracy. Table~\ref{tab:real_results_taw} extends the results presented in the main paper in Table~\ref{tab:real_results} to task-aware measurements. It can be observed that ReReLRP achieves remarkable results on task-aware metrics in the real-life use cases included.

\end{document}